\documentclass[journal, twoside]{IEEEtran}

\usepackage{amsmath,amssymb,amsfonts,bm}
\usepackage{algorithmic}
\usepackage{graphicx}
\usepackage{textcomp}
\usepackage{xcolor}
\usepackage{subfigure}
\usepackage{url,hyperref}

\usepackage{multirow}
\usepackage{makecell}
\usepackage{flushend}
\usepackage{chngcntr}
\usepackage{booktabs,dcolumn}
\newcolumntype{d}[1]{D{.}{.}{#1}} % "decimal" column type
\renewcommand{\ast}{{}^{\textstyle *}} % for raised

\newcommand{\Ll}{\mathcal{L}}
\newcommand{\x}{\bm{x}}
\newcommand{\z}{\bm{z}}
\newcommand{\rep}{\text{rep}}
\newcommand{\aux}{\text{aux}}
\DeclareMathOperator{\Proj}{Proj}
\DeclareMathOperator{\Enc}{Enc}
\DeclareMathOperator{\Clas}{Clas}

\author{Bella Septina Ika Hartanti, Valentino Vito, Aniati Murni Arymurthy, Andie Setiyoko
\thanks{Bella Septina Ika Hartanti is with the Faculty of Computer Science, Universitas Indonesia, Depok, Indonesia (e-mail: bella.septina@ui.ac.id).}
\thanks{Valentino Vito is with the Faculty of Computer Science, Universitas Indonesia, Depok, Indonesia (e-mail: valentino.vito11@ui.ac.id).}
\thanks{Aniati Murni Arymurthy is with the Faculty of Computer Science, Universitas Indonesia, Depok, Indonesia (e-mail: aniati@cs.ui.ac.id).}
\thanks{Andie Setiyoko is with the Remote Sensing Technology and Data Center, Indonesian National Institute of Aeronautics and Space (LAPAN), Jakarta, Indonesia (e-mail: andie.setiyoko@lapan.go.id).}}

\begin{document}

\title{Multimodal SuperCon: Classifier for Drivers of Deforestation in Indonesia}

\maketitle

\begin{abstract}
Deforestation is one of the contributing factors to climate change. Climate change has a serious impact on human life, and it occurs due to emission of greenhouse gases, such as carbon dioxide, to the atmosphere. It is important to know the causes of deforestation for mitigation efforts, but there is a lack of data-driven research studies to predict these deforestation drivers. In this work, we propose a contrastive learning architecture, called Multimodal SuperCon, for classifying drivers of deforestation in Indonesia using satellite images obtained from Landsat 8. Multimodal SuperCon is an architecture which combines contrastive learning and multimodal fusion to handle the available deforestation dataset. Our proposed model outperforms previous work on driver classification, giving a 7\% improvement in accuracy in comparison to a state-of-the-art rotation equivariant model for the same task.
\end{abstract}


\section{Introduction}
Deforestation has become a major problem for the world at large, especially in tropical countries such as Indonesia. Since 2001, the amount of forest area in Indonesia that is overrun by oil palm plantations has doubled, reaching 16.24 Mha in 2019 (64\% industrial; 36\% smallholder). This amount is more than the official estimates of 14.72 Mha \cite{gaveau2022}. At the same time, forest area has declined by 11\% (9.79 Mha), 32\% (3.09 Mha) of which ultimately converted into oil palm plantations, and 29\% (2.85 Mha) were both cleared and converted within the same year \cite{gaveau2022}.

Vegetation and soils in forest ecosystems are crucial in regulating various ecosystem processes \cite{arneth2019framing}. Deforestation, both natural or human-made, can cause droughts due to lack of transpiration in the soil. If this disaster continues and becomes large scale, it will have an adverse impact on the climate. To illustrate, the expansion of oil palm plantations in Kalimantan alone is projected to contribute 18–-22\% of Indonesia’s 2020 carbon emission \cite{carlson2013carbon}. Deforestation also has a destructive impact on the ecosystem and biodiversity, and it is becoming a force of global importance \cite{foley2005}. As one of the countries with the most primary forests in the world, Indonesia has a looming threat of major deforestation. Therefore, analyzing and understanding the drivers for deforestation can help to determine areas in need of reforestation. It also prevents deforestation from becoming more widespread.

Since the advent of high-resolution satellite imaging, disaster analysis typically uses satellite imagery which offers a rich source of information. This can be helpful in determining regions that are currently affected by a disaster, such as deforestation. Prior work have used satellite imagery and deep learning, especially CNN, to solve problems in remote sensing due to their ability to give better performance. In addition, some work have also used multimodal inputs and segmentation to achieve precise predictions of drivers of deforestation in particular. ForestNet \cite{irvin2020} implements scene data augmentation and multimodal fusion \cite{atrey2010multimodal,sheng2020effective} using CNN models. On top of that, another study \cite{mitton2021} introduces the use of rotation-equivariant CNNs on segmentation learning to generate stable segmentation maps with U-Net. However, none of these studies explains in depth the optimal number of auxiliary inputs for the multimodal fusion and the class imbalance naturally present in the dataset.

\par In this work, we propose a deep learning model, called Multimodal SuperCon, which uses contrastive loss and focal loss for classifying drivers of deforestation in Indonesia. The architecture is chosen due to the class imbalance present in the dataset used in \cite{irvin2020}. Using satellite images from Landsat 8, our model consists of two stages of training, namely representation training as the first stage and classifier fine-tuning as the second stage. Our model is a modification of an established architecture known as SuperCon \cite{supercon2022}, and it allows multimodal fusion in the classifier fine-tuning stage.

\section{Related Work}
\subsection{Machine Learning for Deforestation Analysis}
To our knowledge, there have been two proposed deep learning models for handling the deforestation driver classification task in Indonesian forest regions. The first was the ForestNet architecture introduced by Irvin \textit{et al.} \cite{irvin2020} in 2020, and the second was a rotation equivariant CNN architecture applied by Mitton and Murray-Smith \cite{mitton2021} in 2021.

ForestNet \cite{irvin2020} is a semantic segmentation architecture that is created to (1) address that there are often multiple land uses within a single forest loss image, and (2) implicitly utilize the information of specific loss regions. ForestNet allows for predictions from high-resolution (15m) images to predict different drivers of multiple loss regions with varying sizes. The model also incorporates a type of data agumentation named scene data augmentation (SDA), per-pixel classification, and multimodal fusion in its architecture.

Many research studies in deep learning use convolutional neural networks (CNN) due to their powerful capabilities to process sensory data such as images and videos. A CNN model consists of multiple convolutional layers which are translation equivariant. In other words, if an input image is translated to any direction, the result of the feature map is shifted accordingly. However, standard convolutional layers are not rotation equivariant, which can be crucial in order to ensure that the model performs well in certain applications. To make the convolutional layers rotation equivariant, architectures involving group equivariant CNNs can be employed \cite{weiler2019, cohen2016group, cohen2016steerable}. Rotation equivariant CNNs built using this method has been successfully applied in \cite{mitton2021} to the same deforestation dataset used by Irvin \textit{et al.} \cite{irvin2020}.

\subsection{Contrastive Learning}
Contrastive learning is a representation learning approach that aims to shape the feature space in such a way that similar (or positive) samples are clustered together, while dissimilar (or negative) samples are pushed apart. In the past, contrastive learning is executed in a self-supervised manner, with every anchor having exactly one positive sample \cite{chen2020simple, henaff2020, hjelm2018, tian2020}. This positive sample can be obtained by, for example, applying data augmentation to the anchor.

Khosla \textit{et al.} \cite{khosla2020} introduced the concept of supervised contrastive learning, which allows an anchor to have multiple positive samples. The positive samples of an anchor consists of samples that belong to the same class as the anchor. In effect, supervised contrastive learning aims to minimize the distance of intra-class samples and maximize the distance of inter-class samples in the feature space. In practice, supervised contrastive learning is done using the contrastive loss, whose formula is shown in Eq.\ (\ref{eq:contrastive}).

\subsection{Class Imbalance in Datasets}
Class imbalance occurs when the majority of data in a dataset is classified as a certain class. This imbalance may cause difficulty in training the model using data from the minority classes. Several methods have been proposed in the past to deal with class imbalance. One popular solution is by either undersampling the majority class \cite{drummond2003, he2009} or oversampling the minority class \cite{peng2020, shen2016} so that the dataset becomes balanced.

Another way to ensure that the model learns from the minority classes is by modifying the loss function. Despite being the most popular classification loss, the cross-entropy loss is often not the most suitable choice for handling imbalanced datasets. Focal loss \cite{lin2017, yeung2022}, shown in Eq.\ (\ref{eq:focal}), is a widely used modification of cross-entropy loss that is able to focus on hard-to-classify samples. Consequently, focal loss is a more suitable option when dealing with class imbalance. Another loss function that has been successfully utilized is the contrastive loss. A training strategy involving contrastive learning has been shown to be more effective or at least on par compared to cross-entropy and focal loss \cite{marrakchi2021, supercon2022}. 

\section{Methodology}
The model that we apply to the deforestation dataset is built upon the SuperCon architecture \cite{supercon2022}. SuperCon is a two-stage, contrastive learning architecture consisting of a representation training stage and a classifier fine-tuning stage. The resulting model is used for handling classification tasks. During the representation training stage, the ResNet encoder backbone learns from data using the contrastive loss. Afterwards, the classifier fine-tuning stage is used to train a classification layer which generates an estimated probability for each class label. SuperCon works well on datasets under settings involving class imbalance \cite{supercon2022}.

\begin{figure}[!ht]
	\centerline{
	    \includegraphics[width=\columnwidth]{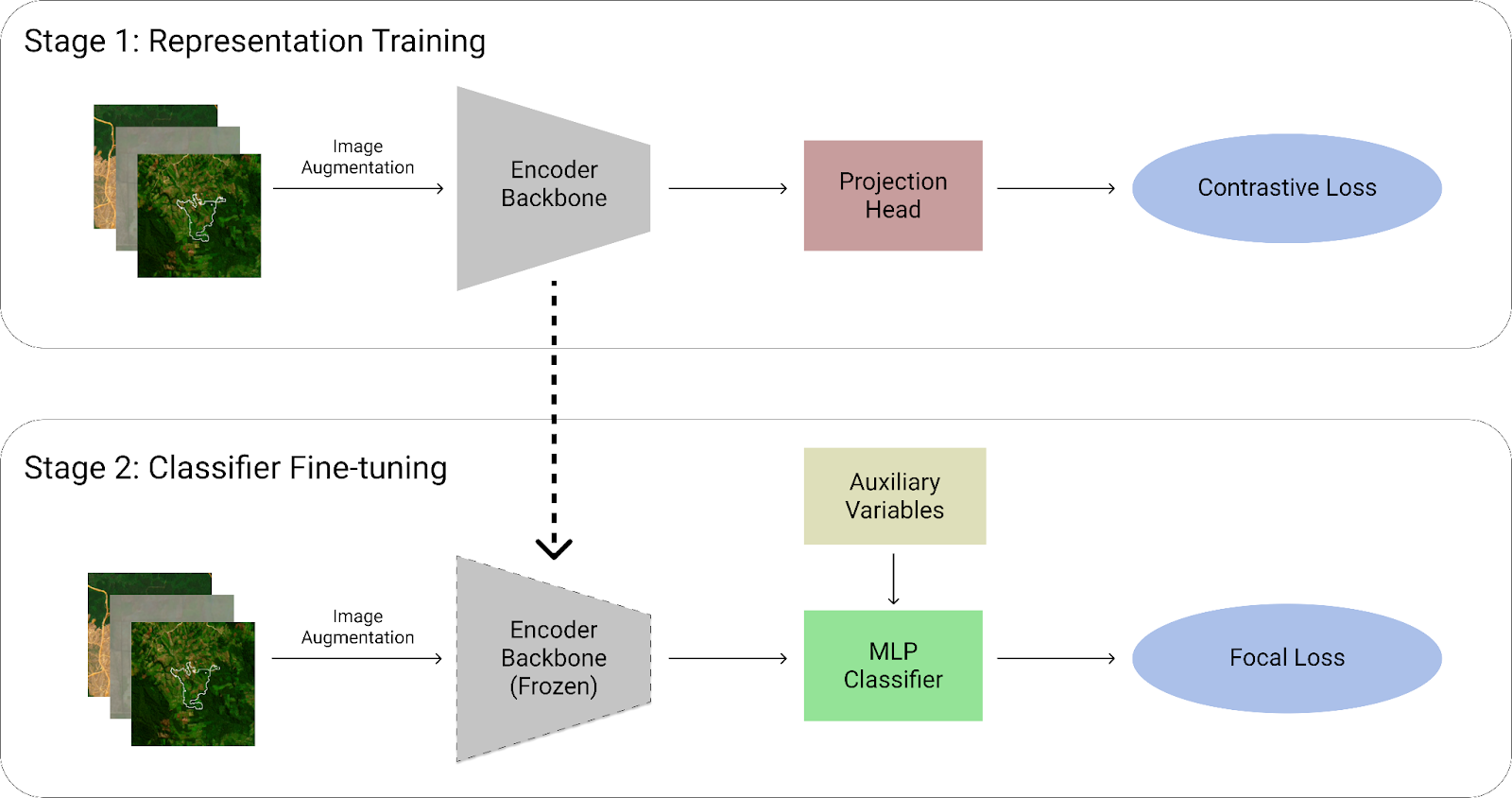}
	}
	\caption {The Architecture of Multimodal SuperCon}
	\label{fig:model}
\end{figure}

We propose Multimodal SuperCon, a modification of SuperCon with the addition of multimodal fusion during classifier fine-tuning. This way, the model can learn from a variety of data in order to obtain a more accurate result. In our application to the deforestation dataset, Multimodal SuperCon allows the use of auxiliary predictors, such as slope, aspect, and elevation, in addition to the satellite images. The architecture of Multimodal SuperCon is illustrated in Fig.\ \ref{fig:model}. The following subsections explain the two stages of Multimodal SuperCon in more detail.

\subsection{Representation Training}
Let $\{\x_1, \dots \x_N\}$ be a mini-batch of augmented training samples from the dataset with batch size $N$. The features of each training sample $\x_i$ are extracted using the encoder backbone $\Enc_\theta$. In practice, the encoder uses a pre-trained model, such as the ResNet or UNet architecture. The encoder $\Enc_\theta$ generates a representation $\rep_i = \Enc_\theta(\x_i)$, which is then projected using the projection head $\Proj$.

The projection head is a two-layer network of size 2048 and 128, respectively. It generates a lower-dimensional representation $\z_i = \Proj(\rep_i)$ of each training sample $\x_i$. Each $\z_i$ is $L_2$-normalized ($\lVert \z_i \rVert_2 = 1$) so that $\z_i$ lies on the unit hypersphere. Afterwards, contrastive loss is applied to the set $\{\z_1, \dots \z_N\}$:
\begin{equation}\label{eq:contrastive}
    \Ll_{\text{con}} = -\sum_{i=1}^n \frac{1}{|P_i|} \sum_{\z_j \in P_i} \log\left(\frac{\exp(\z_i \cdot \z_j/\tau)}{\sum_{\z_k \in A_i} \exp(\z_i \cdot \z_k/\tau)}\right)
\end{equation}
The notation $P_i$ denotes the set of positive samples of the anchor $\z_i$ (that is, samples $\z_j$ in the mini-batch other than $\z_i$ having the same class label as $\z_i$). On the other hand, $A_i$ denotes the set of samples $\z_k$ in the mini-batch other than $\z_i$.

Contrastive loss serves to contrast the training samples among themselves in order to shape the feature space according to class labels. The loss contains the temperature parameter $\tau$. The parameters $\theta$ of the encoder $\Enc_\theta$ are then updated according to the gradient of the contrastive loss:
\begin{equation}
    \theta_{n+1} \gets \theta_n - \lambda \cdot \nabla\Ll_{\text{con}},
\end{equation}
where $n$ is the iteration index and $\lambda$ is the learning rate.

\subsection{Classifier Fine-Tuning}
In the fine-tuning stage, the encoder backbone is frozen so that training is focused on the classifier $\Clas_\varphi$. Each training sample $\x_i$ is fed through the trained encoder to generate $\rep_i = \Enc_\theta(\x_i)$. Afterwards, $\rep_i$ is concatenated with the extracted features $\aux_i$ of the auxiliary variables to generate $\rep'_i = \rep_i \,\Vert\, \aux_i$. This concatenation is the process in which multimodal fusion occurs.

The MLP classifier $\Clas_\varphi$ takes $\rep'_i$ as input to generate an estimated probability for every class. Unlike regular SuperCon, $\Clas_\varphi$ consists of multiple layers instead of just one in order to handle the multimodal fusion. Let $p_t$ be the estimated probability of the ground-truth class label of $\x_i$. The classification loss of the model is calculated using the focal loss:
\begin{equation}\label{eq:focal}
    \Ll_{\text{foc}} = -\alpha_t (1 - p_t)^\gamma \log(p_t)
\end{equation}
The loss $\Ll_{\text{CE}} = \log(p_t)$ is simply the standard cross-entropy loss. Therefore, focal loss is a generalization of cross-entropy loss.

Focal loss contains two parameters: the balancing parameter $\alpha_t$ of the ground truth label and the focusing parameter $\gamma$. Since the encoder is frozen, only the parameters $\varphi$ of the classifier $\Clas_\varphi$ are updated according to the gradient of the focal loss:
\begin{equation}
    \varphi_{n+1} \gets \varphi_n - \lambda \cdot \nabla\Ll_{\text{foc}},
\end{equation}
where $n$ is the iteration index and $\lambda$ is the learning rate.

\section{Experiments}
\subsection{Experimental Datasets}
\begin{figure}[!ht]
	\centerline{
	    \includegraphics[width=0.5\columnwidth]{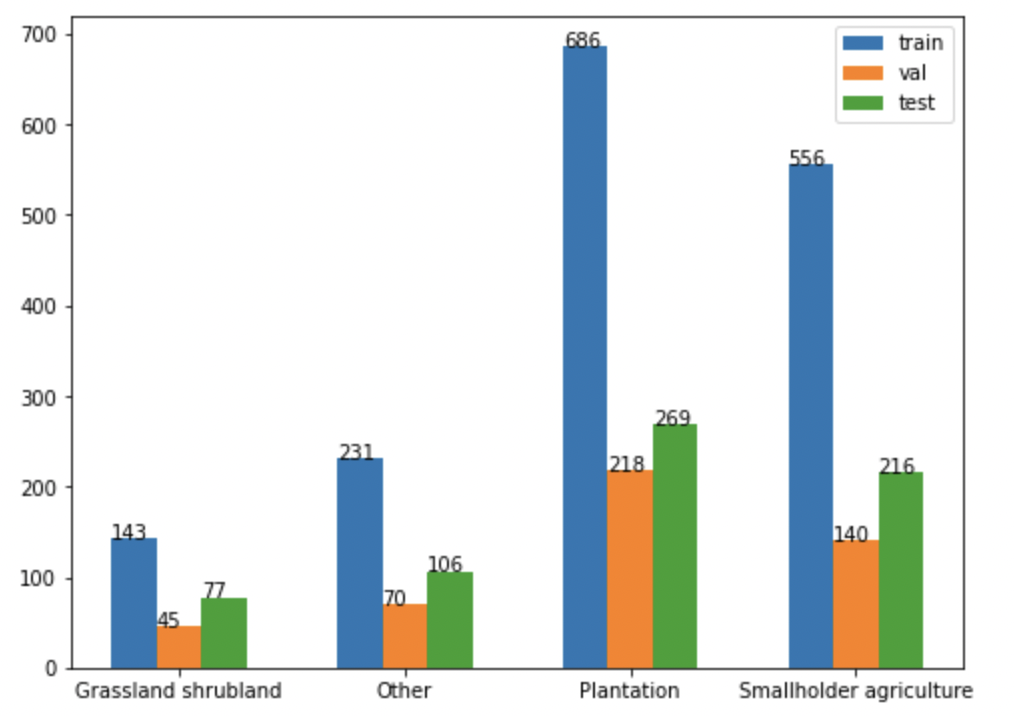}
	}
	\caption {The Distribution of The ForestNet Dataset by Class}
	\label{fig:dataset}
\end{figure}

We use the public Landsat8 image dataset on deforestation in Indonesia to train and test our model. This dataset is the same as the dataset used in \cite{irvin2020, mitton2021}. The driver annotations and coordinates for forest loss events were curated by Austin \textit{et al.} \cite{austin2019}. The dataset consists of 1616 data for training, 473 data for validation, and 668 data for testing. The forest loss images were obtained from various regions no more than five years after deforestation occurred in the area. Each image is classified into one of four classes: grassland/shrubland, plantation, smallholder agriculture, or other. The classes represent the drivers of deforestation of the forest loss region shown in a satellite image. As shown in Fig.\ \ref{fig:dataset}, the data distribution for the four classes is imbalanced, with most of the data belonging to the plantation class.
 
 \begin{figure}[!ht]
	\centerline{
	    \includegraphics[width=0.8\columnwidth]{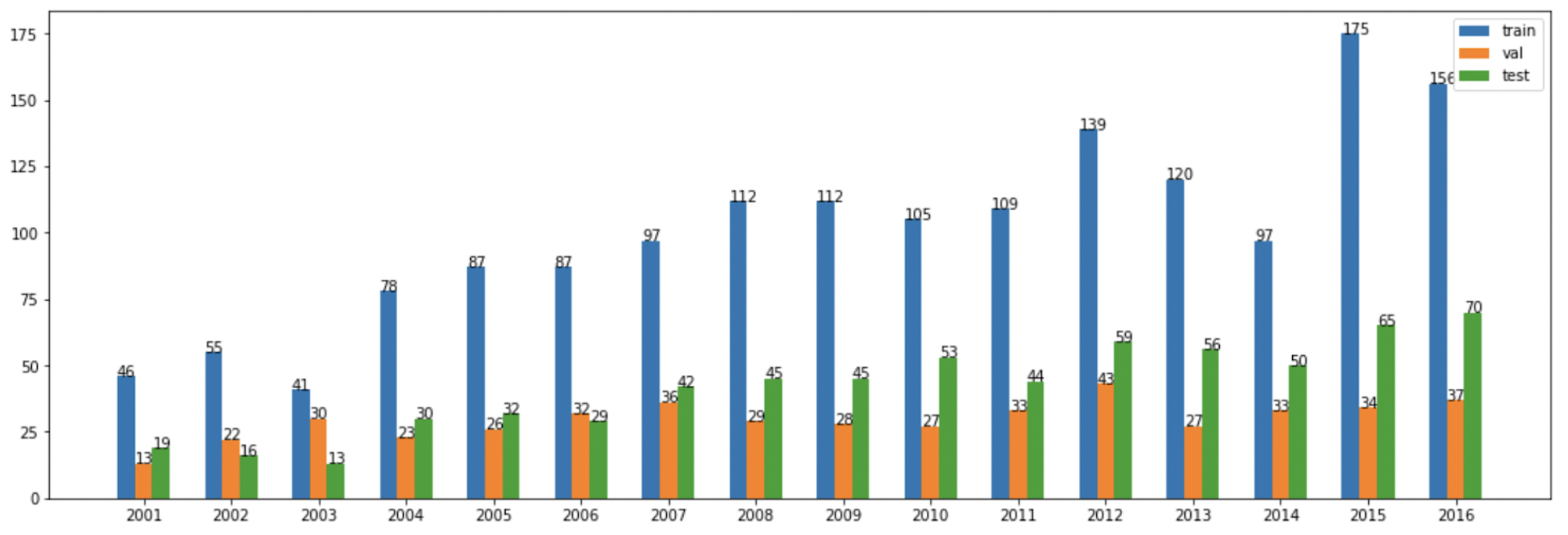}
	}
	\caption {The Distribution of The ForestNet Dataset Based on Year of Occurrence of The Forest Loss Event}
	\label{fig:temporal-aspect}
\end{figure}

\begin{figure}[!ht]
	\centerline{
	    \includegraphics[width=\columnwidth]{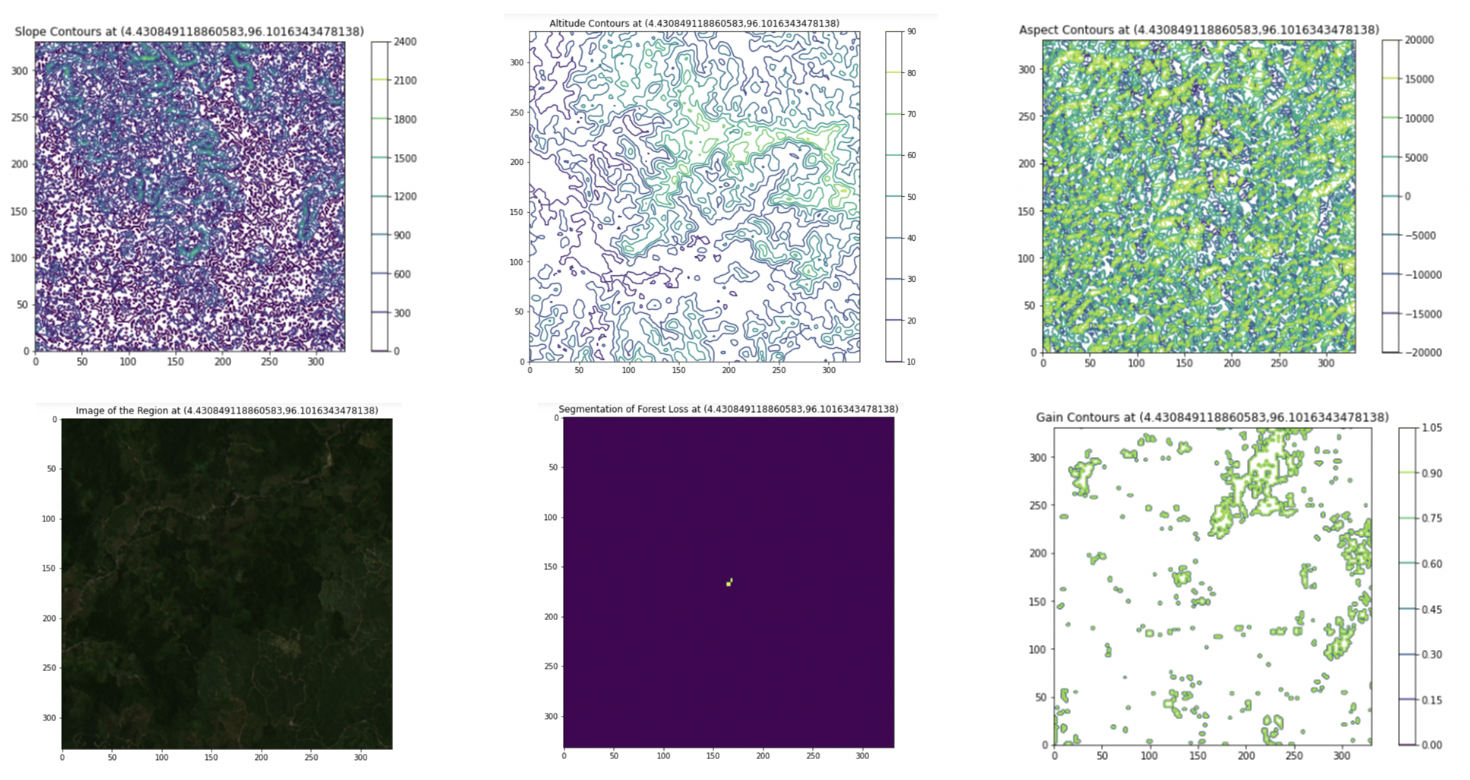}
	}
	\caption {An Illustration of a Landsat 8 Imaging Datum Along With Contour Plots of Its Auxiliary Variables}
	\label{fig:example_data}
\end{figure}

The original dataset is collected from forest loss events at 30m resolution from 2001 to 2016 in various Indonesian islands. Fig.\ \ref{fig:temporal-aspect} shows the distribution of data based on when the forest loss event occurred. Due to the lack of availability of Landsat 8 imagery, which is used for constructing composite images as input data, we only use data obtained in the last five years. Each forest loss region is represented as a polygon, indicating the forest loss event within a year. On the other hand, the composite images are represented as an RGB image with a total size of 332 x 332 pixels to capture the forest loss. We also use auxiliary predictors in addition to the images so that we are able to employ multimodal fusion to the classification task. The auxiliary predictors used in our experiment are slope, altitude, aspect, and gain. Fig.\ \ref{fig:example_data} provides an example of the data used in this research.

\subsection{Experimental Settings}
We use the PyTorch framework to build our Multimodal SuperCon model\footnote{Source code: \url{https://github.com/bellasih/multimodal_supercon}}. We implement several data augmentation techniques, including horizontal flip, rotation, and elastic transform, in order to increase the robustness of the model. The model is trained on NVIDIA DGX A100 with 64 GB memory for roughly 45 minutes. In addition, we use the Adam optimizer and a learning rate of $\lambda = 10^{-3}$ during training. We set aside 15 epochs for the representation training of the encoder backbone and 10 epochs for the classifier fine-tuning. The batch size for both stages is set to 16.

We use EfficientNet-B2 \cite{tan2019} as the encoder backbone for the Multimodal Supercon architecture. For comparison purposes, we also utilize the ResNet18 \cite{he2016deep} and UNet \cite{ronneberger2015} pre-trained models as substitute encoders. We set $\tau = 0.1$ for the contrastive loss, and $\alpha = 0.8$ and $\gamma = 2$ for the focal loss. The architecture is trained a total of three times using only one auxiliary variable (slope), two auxiliary variables (slope and altitude), and all four auxiliary variables (slope, altitude, aspect, and gain), respectively. We compare the performance of our model to the CNN models used in \cite{mitton2021} as baselines.

\subsection{Results and Discussion}
From Tables \ref{tab:rep_learn} and \ref{tab:clas_learn}, the experimental results show that the proposed model (Multimodal SuperCon using EfficientNet-B2 as the encoder and MLP as the classifier) gave the best performance. Compared to other models (ResNet18 as the encoder with MLP; Vanilla U-Net as the encoder with MLP), EfficientNet-B2 excels both during the representation training stage and the classifier fine-tuning stage. The proposed model gave an accuracy of 66\% when using one auxiliary variable (slope) and two auxiliary variables (slope and altitude). An accuracy of 70\% is obtained when using all four auxiliary variables during classifier training. Also, the proposed model gave lower loss on representation learning, namely 0.49 using one auxiliary variable, 0.52 using two auxiliary variables, and 0.48 using four auxiliary variables. These differences indicate that the use of an appropriate pre-trained model is crucial to obtain better performance. EfficientNet outperforms not only in terms of accuracy and loss, but also in terms of computation time.

\par On the optimal number of auxiliary variables used, the best model gave an accuracy of 70\% and a loss of 0.48 when using all four auxiliary variables. This indicates that the addition of several relevant inputs can give more information to the model, hence benefiting training. From Table \ref{tab:comp}, our proposed model also outperforms the results of the baseline model for the same task, namely the rotation equivariant CNN architecture, which yields an accuracy of only 63\% \cite{mitton2021}.

\begin{table}[!ht]
\setlength\tabcolsep{0pt}
\caption{The Performance of Multimodal SuperCon on Representation Learning}
\label{tab:rep_learn}
\begin{tabular*}{\columnwidth}{@{\extracolsep{\fill}} l *{4}{d{2.2}} }
\toprule
 & \multicolumn{4}{c}{Evaluation (Loss)} \\
  &
 \multicolumn{1}{c}{1 Aux} &
 \multicolumn{1}{c}{2 Aux} &
 \multicolumn{1}{c}{4 Aux}\\
\midrule
ResNet18 + MLP  & 0.56 & 0.57 & 0.59 \\
EfficientNet + MLP   & \multicolumn{1}{c}{\textbf{ 0.49}} & \multicolumn{1}{c}{\textbf{ 0.52}} & \multicolumn{1}{c}{\textbf{ 0.48}}  \\
UNet + MLP  & 0.56 & 0.56 & 0.56 \\
\bottomrule
\end{tabular*}
\end{table}

\begin{table}[!ht]
\setlength\tabcolsep{0pt}
\caption{The Performance of Multimodal SuperCon on Classifier Learning}
\label{tab:clas_learn}
\begin{tabular*}{\columnwidth}{@{\extracolsep{\fill}} l *{4}{d{2.2}} }
\toprule
 & \multicolumn{3}{c}{Evaluation (Accuracy)} \\
 & \multicolumn{1}{c}{1 Aux} &
 \multicolumn{1}{c}{2 Aux} &
 \multicolumn{1}{c}{4 Aux}\\
\midrule
ResNet18 + MLP  & 0.47 & 0.55 & 0.51  \\
EfficientNet + MLP   & \multicolumn{1}{c}{\textbf{ 0.66}} & \multicolumn{1}{c}{\textbf{ 0.66}} & \multicolumn{1}{c}{\textbf{ 0.70}}  \\
UNet + MLP  & 0.55 & 0.54 & 0.59 \\
\bottomrule
\end{tabular*}
\end{table}

\begin{table}[!ht]
\setlength\tabcolsep{0pt}
\caption{A Comparison Between the Performance of Multimodal SuperCon and Baseline CNN Models}
\label{tab:comp}
\begin{tabular*}{\columnwidth}{@{\extracolsep{\fill}} l *{2}{d{2.2}} }
\toprule
 & \multicolumn{1}{c}{Test accuracy} \\
\midrule
UNet - CNN \cite{mitton2021} & 0.58\\
UNet - C8 Equivariant \cite{mitton2021} & 0.63\\
Multimodal SuperCon (1 Aux)   & 0.66\\
Multimodal SuperCon (2 Aux)   & 0.66\\
Multimodal SuperCon (4 Aux)   & \multicolumn{1}{c}{\textbf{ 0.70}}\\
\bottomrule
\end{tabular*}
\end{table}

\section{Conclusion and Future Work}
In this paper, we introduced Multimodal SuperCon, a two-stage training architecture involving multimodal fusion to handle the class imbalance present within the deforestation dataset used in \cite{irvin2020,mitton2021}. We applied Multimodal SuperCon to the public Landsat8 imaging dataset consisting of data on forest loss regions in Indonesia. The EfficientNet-B2 + MLP model gave superior performance compared to other encoder architectures, and the resulting test accuracy was significantly higher than the accuracy obtained by the CNN models in \cite{mitton2021}. It was also shown that multimodal fusion is essential in order to yield good results. By using all four auxiliary predictors, the model was able to perform better with an accuracy of 70\%.

In the future, Multimodal SuperCon can be applied to a dataset not necessarily on deforestation datasets, but on any domain requiring some form of multimodal fusion. The model can also be modified for other uses both in deep learning and remote sensing. EfficientNet should be a strong candidate for the encoder backbone since it performs significantly better than other pre-trained models including ResNet18 and UNet.

\section{Acknowledgment}
We thank the support of Tokopedia-UI AI Center of Excellence, Faculty of Computer Science, University of Indonesia, for providing access to the NVIDIA DGX-A100 to run the experiments.

\bibliographystyle{unsrt}
\bibliography{references}

\end{document}